%% file: main.tex
\documentclass[conference]{IEEEtran}
\IEEEoverridecommandlockouts

\usepackage{cite}
\usepackage{amsmath,amssymb,amsfonts}
\usepackage{algorithmic}
\usepackage{graphicx}
\usepackage{textcomp}
\usepackage{xcolor}
\usepackage[normalem]{ulem}
\def\BibTeX{{\rm B\kern-.05em{\sc i\kern-.025em b}\kern-.08em
    T\kern-.1667em\lower.7ex\hbox{E}\kern-.125emX}}

\input{math_defs}
\input{text_defs}

\begin{document}

\title{SiMPLeR: A Series-Elastic Manipulator with Passive Variable Stiffness for Legged Robots}

\author{\IEEEauthorblockN{Sajiv Shah}
\IEEEauthorblockA{
\textit{Saratoga High School}\\
Saratoga, CA. USA \\
sajiv.shah@gmail.com}
\and
\IEEEauthorblockN{Brad Saund}
\IEEEauthorblockA{\textit{Robotics Department} \\
\textit{University of Michigan}\\
Ann Arbor, MI. USA \\
bsaund@umich.edu}
}


\maketitle

\begin{abstract}
We propose a mechanically simple and cheap design for a series elastic actuator with controllable stiffness. Such characteristics are necessary for animals for running, jumping, throwing, and manipulation, yet in robots, variable stiffness actuators are either complicated or mimicked at low bandwidth through feedback controllers. A robust and simple design is needed to build reliable and cheap robots with animal capabilities. The key insight of our design is attaching torsional springs to timing belts to create a variable stiffness linear spring. In an antagonistic pair, varying the distance between motor and joint then varies the actuator stiffness. We build a prototype of our proposed actuator, show the theoretical behavior matches the experimental characterization, and demonstrate an application to robotic one-legged hopping.
\end{abstract}


\input{sections/1_Introduction}

\input{sections/2_Related_Work}

\input{sections/3_Problem_Statement}

\input{sections/4_Method}

\input{sections/5_Experiments}

\input{sections/6_Discussion}

\bibliographystyle{IEEEtran}
\bibliography{references.bib}

\end{document}

%% file: math_defs.tex
\newcommand{\jointtorque}{\tau}
\newcommand{\stiffness}{\sigma}
\newcommand{\groundforce}{F_N}
\newcommand{\leglength}{L}
\newcommand{\shoulderangle}{\theta_1}
\newcommand{\elbowangle}{\theta_2}
\newcommand{\gravity}{g}
\newcommand{\footmass}{m}
\newcommand{\torsionk}{\kappa}
\newcommand{\pretension}{x_i}
\newcommand{\deflection}{\theta}

\newcommand{\springchange}{x_c}
\newcommand{\jointinitial}{\theta_i}
\newcommand{\jointchange}{\theta_d}
\newcommand{\jointfinal}{\theta_f}
\newcommand{\dropheight}{h_1}
\newcommand{\secondheight}{h_2}
\newcommand{\springangle}{\phi}
\newcommand{\stiffnessparameter}{f}
\newcommand{\groundheight}{h_g}

%% file: text_defs.tex
\newcommand{\xxnote}[3]{}
\ifx\hidenotes\undefined
  \usepackage{color}
  \renewcommand{\xxnote}[3]{\color{#2}{#1: #3}}
\fi

\newcommand{\sref}[1]{Section~\ref{#1}}
\newcommand{\figref}[1]{Fig.~\ref{#1}}



%% file: sections/1_Introduction.tex
\section{Introduction}

As robots move from closed-off environments to dynamic environments where they may collaborate with humans or interact with unknown surfaces, many dangers arise \cite{Zinn2004playing}.
With the use of compliant actuators, the force and energy of a collision are absorbed by the actuator itself and the harm on the external object is minimized \cite{an2019mechanical}. 
Compliance has various other benefits, such as energy-efficient locomotion, as observed in human joints \cite{Hu2014compliance}.
Compliance is implemented in robots through two major methods, passive and active compliance. 
Active compliance uses a stiff actuator with integrated software control to create a virtual spring \cite{Villani2016force}, while passive compliance uses a physical elastic component between the motor and output \cite{Ham2009compliant}. 
While active compliance is more mechanically simple and allows for simple stiffness variation, passive compliance reacts much faster and can store energy. 

\begin{figure}
    \centering
    \includegraphics[width=\linewidth]{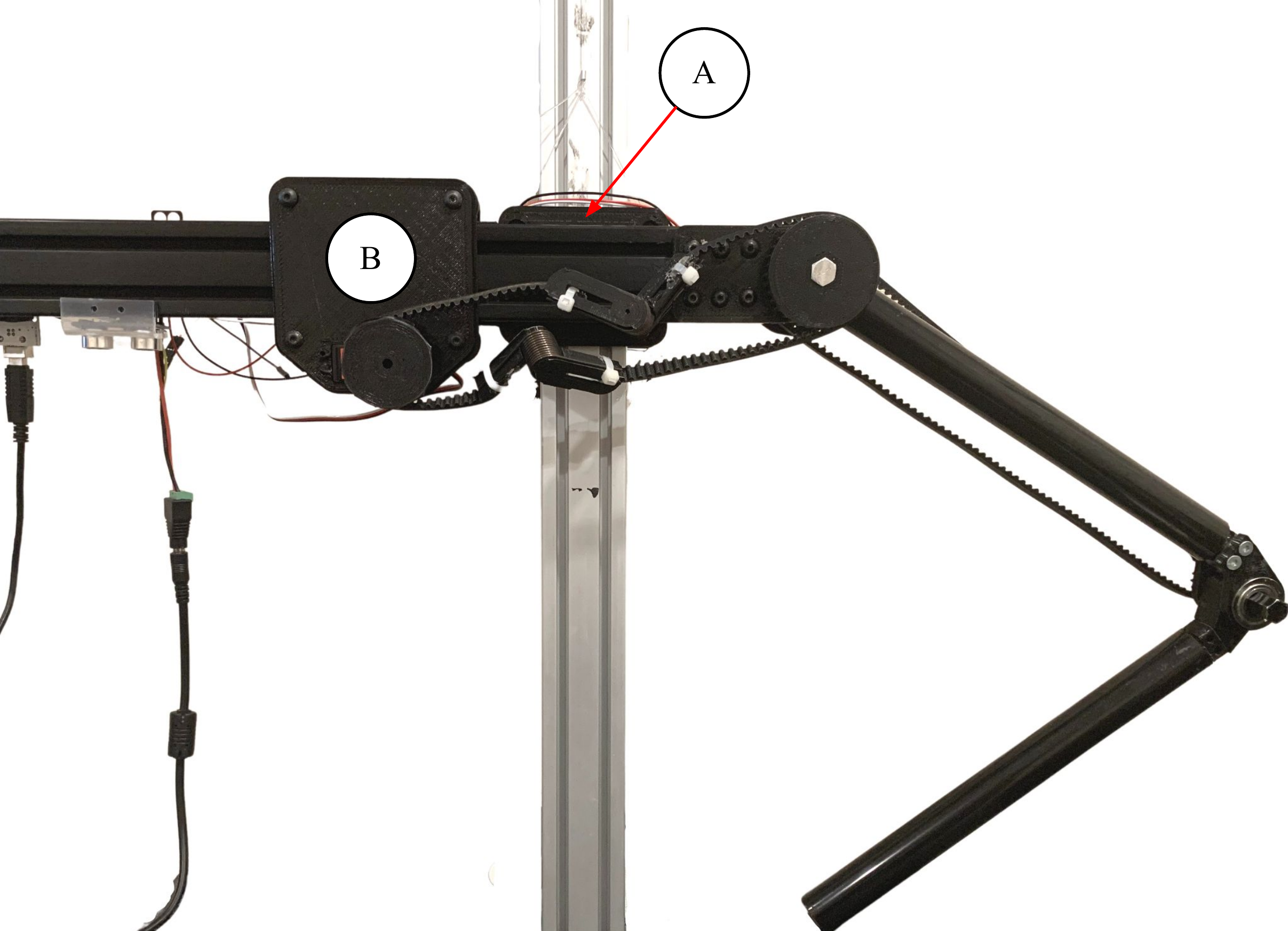}
    \caption{Application of our simplistic variable stiffness actuator to a single-legged robot.}
    \label{fig:realImage}
\end{figure}

Fixed compliance or stiffness is extremely limiting, especially in legged robots. 
Animals and Humans require variable stiffness in their legs to achieve stable running \cite{rummel2008stable}.
To achieve higher frequency running for the same range of motion, the stiffness of the leg must increase as well \cite{farley1991hopping,mutlu2018effects}. 
Fixed passive compliance limits the maximum operation speed of robots and the payload they can carry, as any load that would cause a deflection too great by the elastic components can result in robot failure.
Robots with fixed stiffness cannot carry out various tasks, and are therefore far less desirable than those with variable stiffness \cite{galloway2010}.
Passive Variable Stiffness Actuators (VSAs) address these problems by allowing robots to vary the compliance in their joints \cite{wolf2015variable}.
While compliant mechanisms apply a restoring force that is proportional to their elastic deflection, VSAs decouple this nature and use an additional actuator to vary stiffness. 

Our design addresses the complexity and cost of existing VSAs by using low-cost torsion springs. 
We configure these torsion springs, which have a linear torque-angle relationship, to apply a non-linear inward force in a single dimension.
By using two of these springs in a timing belt transmission, we create an antagonistic setup similar to that of animal muscle-pairs.
We adjust the pretension of our springs with an additional actuator which simply adjusts the distance of the drive motor relative to the joint, pulling or relaxing the springs.

Our key contribution is a simple, low-cost design with a high amount of variability for applications to various robots.
Our design allows our actuator to be applied into many robots that use belt transmissions, and requires only the addition of the torsion springs and an actuator for joint-motor separation.

In this paper, we present the design, analysis, application, and experimentation of a simplistic variable stiffness actuator. The structure of this paper is as follows. In section II, we explore the existing designs of VSAs and evaluate them. In section III, we define our problem. In section IV, we present and analyze the design of our actuator. In section V we evaluate the performance of our design when applied to a robotic leg. And lastly, in section VI, we draw conclusions from our work.

%% file: sections/2_Related_Work.tex
\section{Related Work}

While many commercial robots such as ASIMO \cite{hurst2008role} use active VSAs as they are mechanically simpler and now feasible due to improvements in electronics and software, passive VSAs are much faster and are more suitable for environments in which robots experience sudden impacts or collisions. Due to these benefits, we mainly cover the work of passive VSAs.
Many passive VSAs are modifications of concepts brought about by fixed compliant actuators. Methods of achieving passive variable stiffness fall into three main categories: Equilibrium controlled stiffness, Antagonistic controlled stiffness, and structure controlled stiffness.

Equilibrium controlled actuators simulate variable stiffness with impedance control.
The traditional series elastic actuator (SEA) \cite{Pratt1995series} uses linear springs to create physical compliance and sensors to measure joint deflection, from which the net torque is inferred.
Control of the motor adjusts the equilibrium position of the system and can simulate a range of effective stiffnesses. 
Robots such as the Kuka iiwa and Franka arms use such actuators with a very stiff elastic coupling and high frequency control.
While this actuator has the advantage of only requiring a single motor for both stiffness and position control, it is does not perform well at the extremes of pure position or pure stiffness control. In the case of a sudden impact, errors due to controller bandwidth are likely. Precise position control is difficult even when moving slow, as slight oscillation from the springs cannot be controlled.

Antagonistic controlled stiffness solves these issues by modifying the SEA to use non-linear springs and a pair motors for stiffness and position control \cite{English1999mechanics}. 
Tonietti \cite{Tonietti2005design} and Migloire  \cite{Migliore2005biologically} use a human-muscle inspired setup, which consists of two motors connected to non-linear springs which lie on either side of an output pulley. The motors stress and relax the springs.
When both motors turn in the same direction, the equilibrium position of the output moves, but if they rotate in different directions, the stiffness of the output either increases or decreases. 
This allows for much better control of both stiffness and control of the output.
This design is also used in the modular VSA-Cube servo actuator \cite{catalano2011VSAcube}.
Despite the improvements in control, these actuators endure high energy cost from spring extension, and are complex in structure due to the complex implementation of non-linear springs.

The Actuator with Mechanically Adjustable Series Compliance (AMASC)\cite{Hurst2004AMASC} mechanically decouples stiffness and position control, as each motor is assigned with exclusively controlling either position or stiffness. This allows for optimization in motor selection, as the two operations have different control and power requirements, which can now be accounted for. However, the design of the actuator itself is complex, involving a large system of pulley and belt transmissions. It has demanding space and component requirements for reliable usage. 

An alternate approach categorized under the antagonistic controlled stiffness method is to use pneumatics as elastic components instead springs. When in a non-rigid container, air is naturally compliant as it is compressible. Pneumatic Artificial Muscles (PAMs) such as the popular McKibben design take advantage of this to use them as non-linear springs \cite{chou1996measurement}. The McKibben muscle expands radially and shrinks axially as it is pressurized, creating a pulling force at its ends. Using these muscles in an antagonistic fashion allows for variable stiffness, but actuation is difficult to control, pressurization is slow, and the infrastructure for on-board pneumatic systems is demanding. Raibert demonstrated this with the pneumatic telescoping-leg robot, which uses hydraulic actuators to pressurize the legs \cite{raibert1989dynamically}. A tethered supply had to be used due to the large air requirements, making the system infeasible for legged robot applications.

Instead of using a pair of non-linear springs, structure controlled stiffness explores the variation of the structure of materials to control stiffness. One method of doing so is changing the effective length of a spring by making some coils inactive \cite{Hollander2005Jack}. Another is to change the effective length of a leaf spring by holding down a section using a motorized slider \cite{Morita1995DesignAD}. Both systems have high friction and high load bearing. The leaf spring actuator uses a large amount of space in various stiffness settings. A far more compact mechanism was developed, which changes the moment of an inertia of a beam inside a helical spring to manipulate the effective stiffness in a specified axis \cite{hollander2004concepts}. However, this mechanism is not able to limit the passive compliance of the joint to a single axis of rotation, and the spring experiences buckling past a small deflection. 

Multiple biped robots that have been recently developed employ the MACCEPA actuator\cite{geeroms2018energetic}. The MACCEPA actuator is based of a design that changes the linkage point of a linear spring on two lever arms \cite{duindam2005optimization}. While this actuator requires 3 motors, 2 for the sliding along the lever arms and 1 for the joint rotation, newer versions only require two. The MACCEPA actuator can be made using standard off the shelf components, but takes up a large amount of space and therefore is only deemed useful in robots that have long lever arms such as robotic arms.

%% file: sections/3_Problem_Statement.tex
\section{Problem Statement} \label{section:Problem}

We desire a compliant actuator composed of a controllable motor angle and an elastic element linking the motor and joint with \textit{controllable stiffness}. 
Specifically, consider the torque applied by the actuator on the joint $\jointtorque(\deflection, \stiffnessparameter)$, where $\theta$ is the relative angular displacement of the motor and joint, and $\stiffnessparameter$ is a controllable parameter that varies the stiffness.
While $\deflection$ can be controlled \textit{at low frequency} by varying the motor angle, $\deflection$ cannot be controlled with high enough rate to control $\jointtorque$ from a sudden change in joint angle, such as an impact.
By varying stiffness through controlling $\stiffnessparameter$ the torque response can be adjusted without the need for high frequency control of $\deflection$.
As a further constraint, the stiffness should be passively held, and thus not require energy to maintain the same stiffness.

A traditional SEA offers only a controllable motor angle with a \textit{fixed} stiffness linkage to the joint.
Our problem requires an SEA with the additional ability to control the stiffness.
As we seek to use cheap and common materials, the stiffness variation will be accomplished through mechanical linkages as opposed to novel material variation.
Thus, the design will require two actuators: one for controlling the motor joint angle, and one for adjusting the stiffness.
As we will discuss in section \ref{section:methodA}, the naive approach of pretension applied to linear springs does not achieve our goal.


We aim to apply this actuator to a legged robot to achieve oscillation-less landings from various drop heights. We define the drop height $h_1$ as the height from the initial drop position to the first contact of the leg with the ground (\figref{fig:legdrop}). For legs with compliance, the springs must contract and expand to absorb the energy of the drop, as well as to apply a torque to balance the weight of the leg. This causes an additional joint deflection $\jointchange$ and an additional change in height $h_2$ given by:

\begin{figure}
    \centering
    \includegraphics[width=0.9\linewidth]{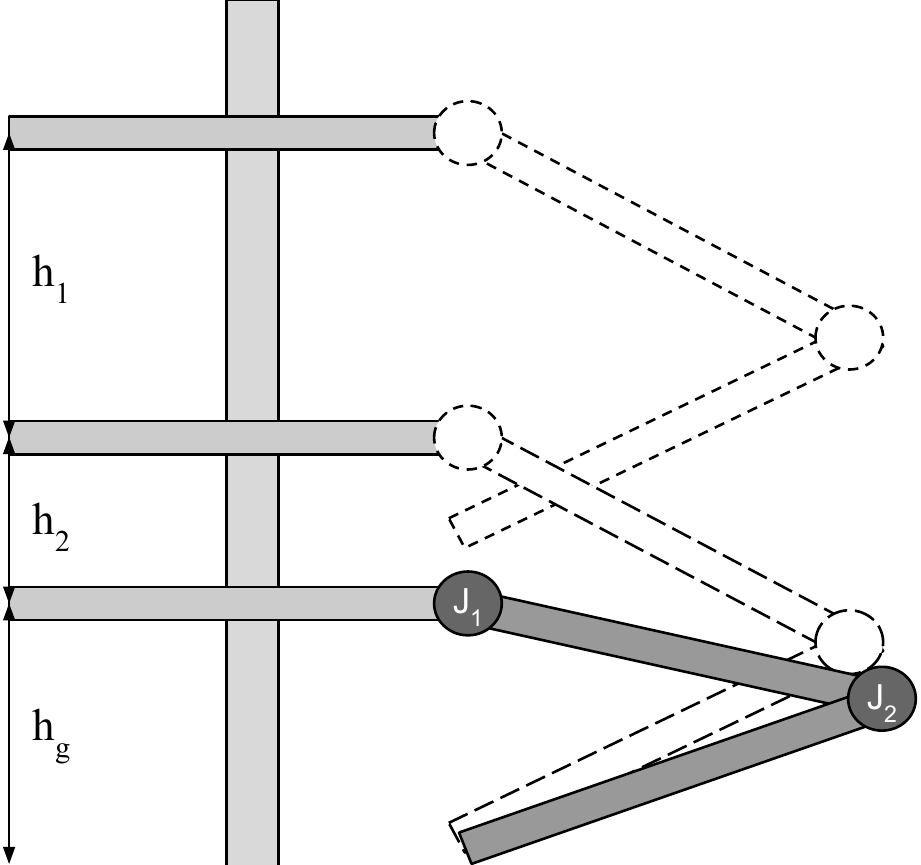}
    \caption{Diagram of leg dynamics in landing. The actuator begins at a height $\dropheight$ that represents the distance to the point of first contact. The leg then drops another distance $h_2$ until the robot is resting at a final height $\groundheight$ above the ground. }
    \label{fig:legdrop}
\end{figure}

\begin{align}
    \jointfinal &= \jointinitial - \jointchange \\
    h_2 &= 2\leglength[\sin{\jointinitial} - \sin{\jointfinal}]
\end{align}
where $\jointinitial$ is the angle of the leg while dropping $h_1$, and $\leglength$ is the length of one leg segment.

An oscillation-less drop requires the drop energy to be completely absorbed by the springs. For a fixed value for parameter $\stiffnessparameter_0$ an oscillation-less drop satisfies:
\begin{align}
    mg(h_1 + h_2) &= C\int_{\theta_i}^{\theta_i+ \theta_d} \jointtorque(\deflection, \stiffnessparameter_0) \ d\theta \label{eq:fixedDrop}
\end{align}
where $m$ is the mass of the system, and $C$ is any constant such as the radius of the pulley the springs act upon.

While $h_2$ is bound by the height of the overall robot, $h_1$ is limited only by the environment the robot acts in, as it may be necessary to drop from various heights to navigate terrain. With a fixed compliance, oscillation-less dropping is not achieved for a large range of drop heights. If $h_1$ is large relative to the stiffness constant $k$, then the springs are not able to absorb all the energy without joint deflection larger than mechanically possible, causing the robot to bottom out. In the opposite case, energy is abruptly stored in the springs, and there will exist a net torque about the leg, causing oscillation.

In our work, we assume the leg segments to have no mass, and energy to be transferred from gravitational potential to spring potential without loss. We additionally do not consider any damping from the landing surface or other internal factors.  We evaluate the effectiveness of our actuator through the achieved stiffness range, the ability to act analogous to a completely rigid actuator, and the performance in application to legged robot functions.

%% file: sections/4_Method.tex
\section{Method}

\subsection{Mechanical Design} \label{section:methodA}
We propose a simple mechanical actuator design to allow control and adjustment of performance using replaceable components.
Our design is based on a modified version of a SEA. 
SEAs are mechanisms that utilize an elastic element in the transmission between the motor and joint to generate mechanical compliance.
When elastics are configured antagonistically in a belt transmission (\figref{SEAdrawing}) , a displacement $\theta$ of the joint relative to the motor results in a net torque towards the rest angle ($\theta$=0).
If linear elastics are used, then the torque on the joint is given by:

\begin{figure}
    \centering
    \includegraphics[width=\linewidth]{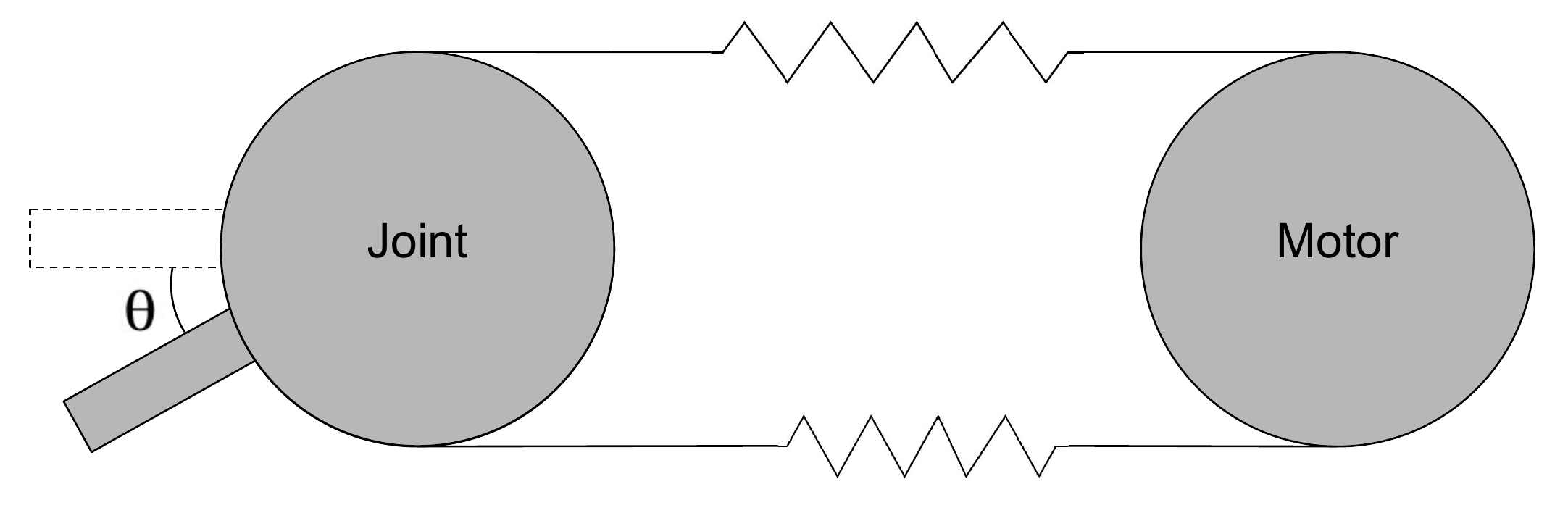}
    \caption{Drawing of Series Elastic Actuator with linear springs.}
    \label{SEAdrawing}
\end{figure}

\begin{align}
    \jointtorque(\theta) &= k[(x_i + r\theta) - (x_i - r\theta)] \label{SEAtorque1}\\
    \jointtorque(\theta) &= k[2r\theta] \label{SEAtorque2}
\end{align}
where $r$ is the radius of the joint pulley, $k$ is the stiffness constant of the elastics, $\tau$ is the torque on the joint, and $x_i$ is the initial length of the springs.
This mechanism has a fixed compliance as the torque created by the springs to restore the joint position cannot be varied independently of the angle the joint is displaced. 
In particular, stretching the top and bottom springs simultaneously (e.g. by changing the distance between the joint an motor) changes each spring force, but does not change the net torque $\jointtorque$.


To achieve our objective of a variable stiffness actuator, we design non-linear elasticity in the antagonistic springs.
This allows controlled variation in the instantaneous spring stiffness $k$ through intentional variation of the initial joint-motor separation $x_i$, which is the controllable parameter $\stiffnessparameter$ defined in \sref{section:Problem}.
We replace ordinary springs in a SEA with off-the-shelf torsional springs in a linear fashion to achieve stiffness dependent on the pretension of the springs.


Torsion springs apply a torque along their legs based on the angle between their legs. Similar to linear extension or compression springs, the torque applied can be approximated by a fixed torsional stiffness constant $\torsionk$ multiplied by the angle of displacement $\springangle$ 
However, when pulled apart outwards by their ends linearly, torsion springs supply an inward force that has non-linear relationship with the linear extension $x$,
which is half the joint-motor separation distance $\frac{\pretension}{2}$ (Fig. \ref{fig:linearForceDiagram}). 
Since the stress-strain relationship is non-linear, there is no longer a spring ``constant". We refer to the instantaneous change in stress per change in strain as the stiffness function $\stiffness$:

\begin{figure}
    \centering
    \includegraphics[width=0.8\linewidth]{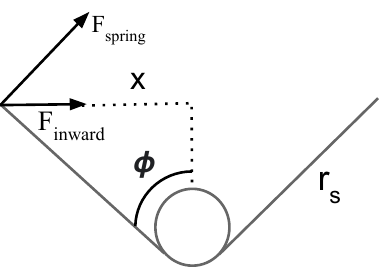}
    \caption{Force diagram of torsion spring. 
    }
    \label{fig:linearForceDiagram}
\end{figure}

\begin{align}
    \stiffness(x) = \frac{ \torsionk \cdot \arcsin(\frac{x}{r_s})}{x \cdot  \sqrt{r_s^2 - x^2}}
    \label{eq:stiffness}
\end{align}
where $r_s$ is the length of a leg of the torsion spring, in meters. The linear extension reaches a maximum at $r_s$, as the legs of the spring are 180 degrees apart. At this point this spring acts analogous to a rigid body in the dimension parallel to its legs, as the spring force acts perpendicular to this dimension. 

We replace the linear springs in a traditional antagonistic belt-driven SEA with our torsion springs, and determine that stiffness can be varied by changing the initial length of the torsion springs, which determines their pretension. To automate the process of setting the desired pretension, an additional actuator is required. To maintain the extension of the springs, a high energy cost is endured if the extension actuator is backdrivable. Additionally, in the case of power failure while the actuator is carrying a load, the springs would release their energy and collapse, creating a low joint stiffness. This could result in catastrophic hardware failure, as a robot such as a quadruped will collapse to the ground.  When using a mechanism such as an inefficient acme screw or a worm gear, which both have a low tendency to backdrive, the energy cost decreases and the system holds its position in case of a power failure. 

\begin{figure}
    \centering
    \includegraphics[width=\linewidth]{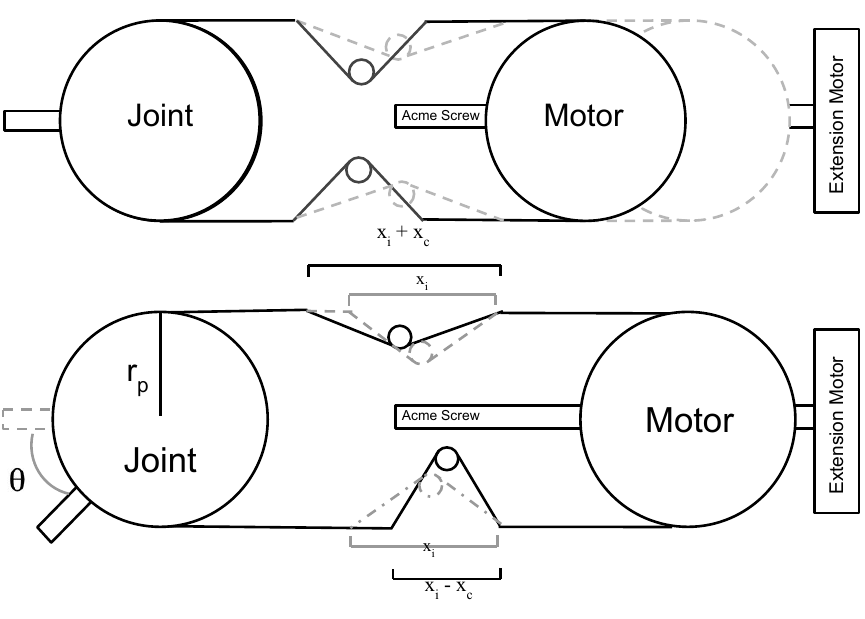}
    \caption{Example of decreasing pretension by using an acme screw to move the motor closer to the joint (top). Demonstration of torsion spring dynamics as the joint is deflected independent of motor, causing the torsion springs to either contract or expand by a length $x_c$ from their initial length, $x_i$.
    }
    \label{fig: acmeExtension}
\end{figure}

\figref{fig: acmeExtension} demonstrates how an acme screw is used in our actuator to move the motor location relative to the joint, which in turn changes the pretension of the springs. It also shows the dynamics of the torsion springs as the joint is displaced independent of the motor. In our antagonistic setup, the torque on the joint when the spring forces are imbalanced is a function of the pretension $x_i$ and the angle of displacement $\theta$, and is modeled by the equation:

\begin{align}
    \jointtorque(\theta, x_i) &= r_p[(x_i + r_p\theta)\stiffness(x_i + r_p\theta) - (x_i - r\theta)\stiffness(x_i - r\theta)] 
    \label{eq:VSAtorque}
\end{align}
 where $r_p$ is the radius of the pulley and $\sigma(x)$ is the stiffness function defined in Eq. \ref{eq:stiffness}. Due to space requirements, $r_p$ must at a minimum be $r_s$ in order to operate in the full stiffness range, or the springs will rub against the belt in low stiffness settings.

\begin{figure}
    \centering
    \includegraphics[width=\linewidth]{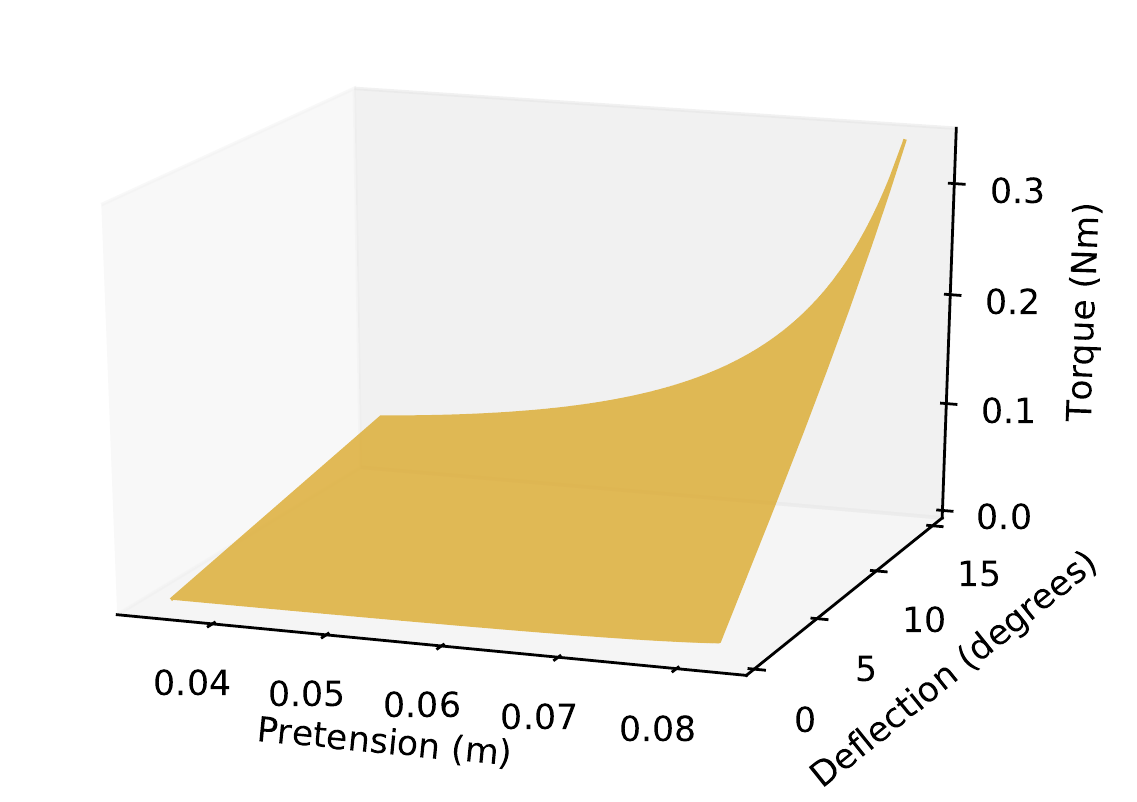}
    \caption{Plot of joint torque for an actuator with torsion springs of rotational stiffness $\kappa = 0.5 Nm/\theta$ and leg length $r_s = 0.05$m. Pretension extension range of $0.035$m to $0.0825$m is shown to demonstrate the general relationship between pretension, deflection, and torque. At pretension extremes, variances are produced.}
    \label{fig:3dplot}
\end{figure}

Fig. \ref{fig:3dplot} shows a plot of the general relationship between the pretension, deflection, and torque. The actuator experiences the biggest variance in joint torque in the upper range of pretension, and a low variance in joint torque in the lower range of pretension.

At a high pretension, the deflection angle range is decreased a large amount as the torsion spring can only extend to a maximum length of $2r_s$. 
At the maximum pretension where $ \pretension = 2r_s $, compliance is theoretically non-existent, but in practice there exists a spring from the torsion spring body and the elasticity of the belt. 
At a low pretension, the deflection range is also limited as the contracting spring may contract to a length of nearly 0, meaning the spring no longer applies any force which causes the timing belt to skip on the output pulley. 



\subsection{Application to Legged Robot}


Variable stiffness is desirable in legged robots for a multitude of reasons. When in motion, for a given leg velocity, touchdown angle, and terrain, there exists an optimal leg stiffness which would result in the most energy efficient operation. For example, when interacting with terrain that is compliant, the surface acts as a spring on the leg. This affects the total joint stiffness, and varying terrain will produce varying behavior on an actuator which has fixed compliance. 

Additionally, legged robots with fixed compliance cannot precisely land from various jump heights without oscillating. To achieve an oscillation-less drop, the energy of the leg must be captured in the springs of the actuator at the desired landing height, and the net torque created by the spring pair contraction and expansion must balance the weight of the robot. 
For a joint with fixed stiffness $k$, equation \eqref{eq:fixedDrop} demonstrates that oscillation-less drop is only achievable for a small range of heights through the variation of the initial leg angle.

With variable-stiffness, we can adjust the pretension to achieve both these things. In \sref{section:Problem} and \figref{fig:legdrop} we explain the dynamics of landing for complaint robot legs. With variable-stiffness, the energy capture of the drop by the springs is modeled by:

\begin{align}
    mg(h_1 + h_2) &= \int_{\pretension}^{\pretension + \springchange} x[\stiffness(x)] dx - \int_{\pretension - \springchange}^{\pretension} x[\stiffness(x)] dx
\end{align}
where $x_c$ is the change in spring length upon landing and approximated by $x_c = r\jointchange$.

At the resting position where the springs have experienced co-contraction and captured the energy drop, the net force on the leg system must also be 0 for there to be no oscillation. For this to be true, the torque about the thigh and knee induced by gravity and the spring deflection must balance.
The torque the co-contraction of springs creates on the thigh is modeled by \eqref{eq:VSAtorque}.
The torque about the thigh $\jointtorque_1$ and the knee $\jointtorque_2$ are approximated by:

\begin{figure}
    \centering
    \includegraphics[width=0.8\linewidth]{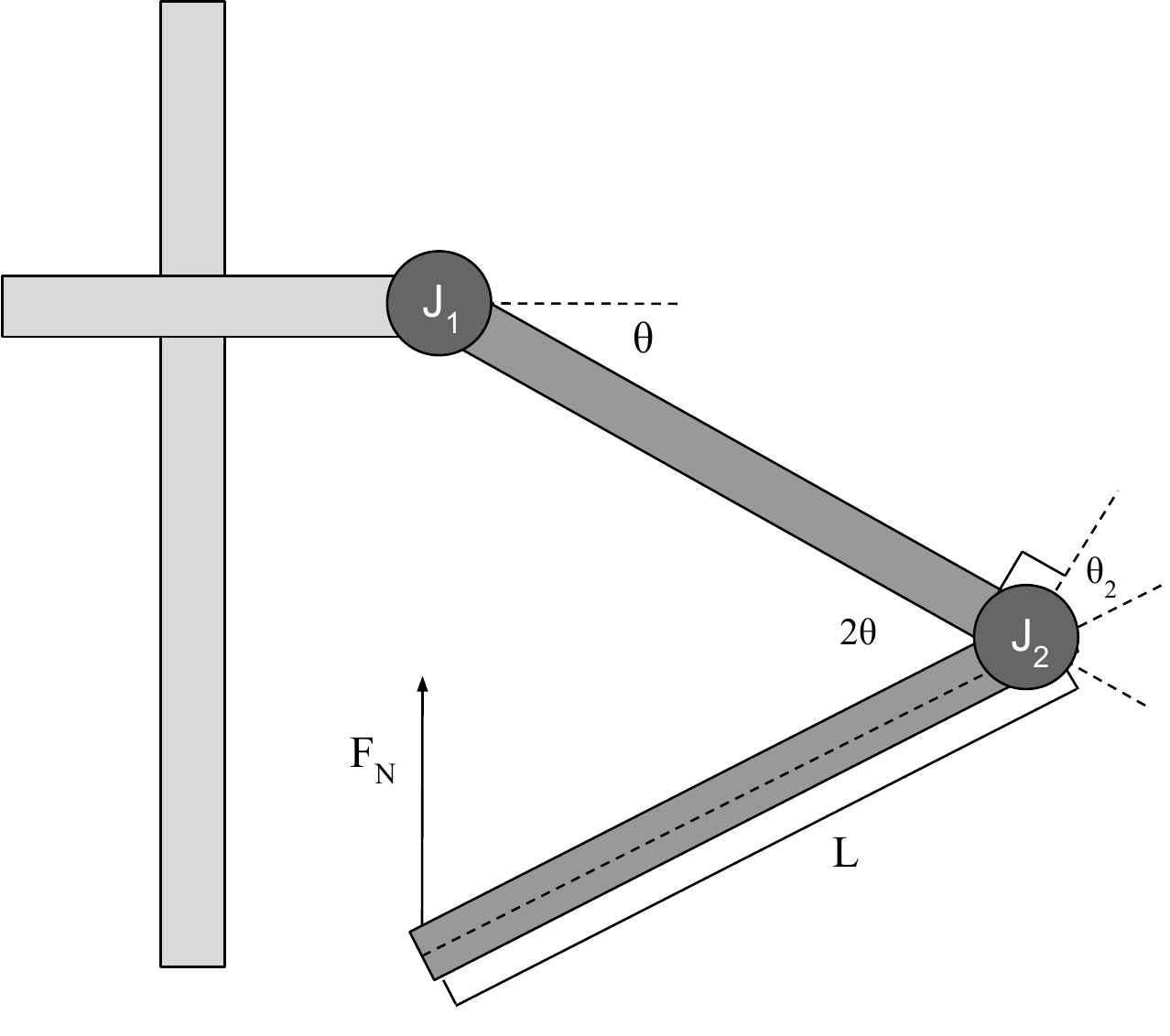}
    \caption{Diagram of forces on leg system when at rest on ground.}
    \label{fig:legtorque}
\end{figure}

\begin{align}
    \groundforce &= \footmass\gravity \\
    \jointtorque_1 &= \groundforce \leglength \cos(\frac{\pi}{2}-\shoulderangle)\cos(\elbowangle)\\
    \jointtorque_2 &= \groundforce \leglength \cos(\shoulderangle) 
\end{align}
where $\shoulderangle$ is the angle of the upper leg segment relative to the horizontal, and $\elbowangle$ is the angle between the leg segments past the perpendicular point(\figref{fig:legtorque}). These angles change as the leg drops a height $h_2$ and the springs contract or expand by $x_c$.
For a given set of parameters ($\groundheight, \dropheight + \secondheight, \jointinitial$), oscillation-less drop is achieved for a single pretension value $\pretension$.

While we experiment with the impacts of variable-stiffness for oscillation-less landing, legged robots more importantly benefit from variable stiffness to maintain structural integrity in landing situations. From large drop heights, overly stiff legs will place high stresses on the joints, and overly soft legs will cause the robot to bottom out, as a large spring extension is required to absorb the leg energy, which causes a large joint deflection. 
This causes potential hardware damage to the robot and makes locomotion impossible.

Tunable stiffness also allows a single robot to efficiently perform various tasks such as work output (running and hopping) in which the passive elastic components can recycle energy, and high power tasks in which a burst of energy must be released or captured (jumping and landing). 

We designed a single leg system to which we implemented our simple variable stiffness actuator to evaluate its impacts and usability. In \sref{section:Experiments} we present our system, the experiments we designed, and our results.

%% file: sections/5_Experiments.tex
\section{Experiments} \label{section:Experiments}

\subsection{Leg Design}

Our leg (\figref{fig:realImage}) consists of a upper and lower segment, which rotate about a thigh and knee joint, respectively. The thigh joint is mounted to a horizontal piece of extrusion. On the extrusion, two sliding mechanisms are mounted, one which is not motorized and allows the leg to slide up and down a on a vertical beam (label A), and one that allows the thigh motor to change its distance relative to the joint (label B).

\begin{figure}
    \centering
    \includegraphics[width=\linewidth]{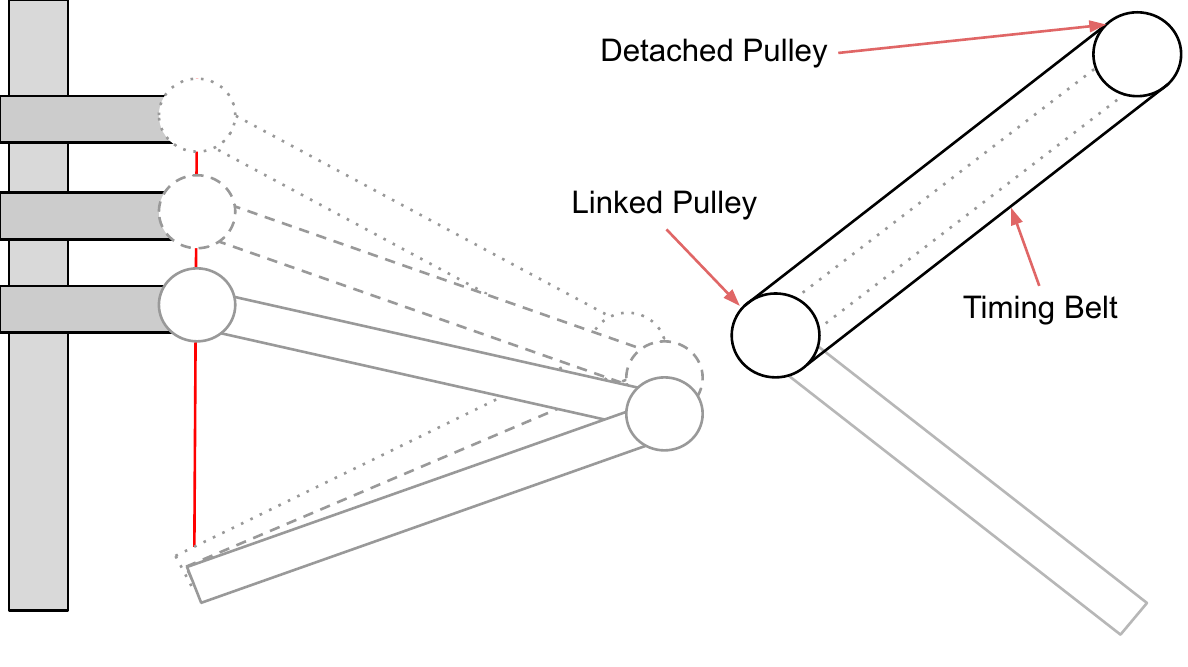}
    \caption{The cardan gear system we use allows us to move the leg with a single actuator such that it travels along a vertical line (shown in red on the left). The detached pulley is mounted concentric to the thigh joint, and is fixed to the body of the robot but not connected to the upper leg segment. The linked pulley is mounted on the knee joint and is linked through an axle to the lower leg segment.}
    \label{fig:cardan}
\end{figure}

To accurately test the use of our actuator, we mechanically linked the knee joint to the thigh joint using a cardan gear system \cite{daskalov1990kinematic}.
This system rotates the knee joint at twice the angle of the thigh joint rotation, keeping the foot, or contact point of the leg at the ground, completely vertical with the point at which the thigh joint mounts on the horizontal beam (\figref{fig:cardan}).
Rotating the thigh joint changes the vertical distance between these points, and therefore the height of the system.

To experiment with the functionality of our spring-system design, we equipped the leg with multiple sensors. We use an absolute encoder that mounts on the thigh joint axle to measure angular displacement, and an ultrasonic distance sensor mounted to the bottom face of the horizontal extrusion to measure the leg height. On the end of the extrusion a servo motor is fixed and is attached to a slide using fishing line. On the slide is mounted a second servo motor, which hosts the actuation mechanism of the thigh joint. All of these systems are powered by a 5v power-supply, and controlled by an Arduino Uno.

\subsection{Characterization} \label{section: Characterization}

\begin{figure}
    \centering
    \includegraphics[width=\linewidth]{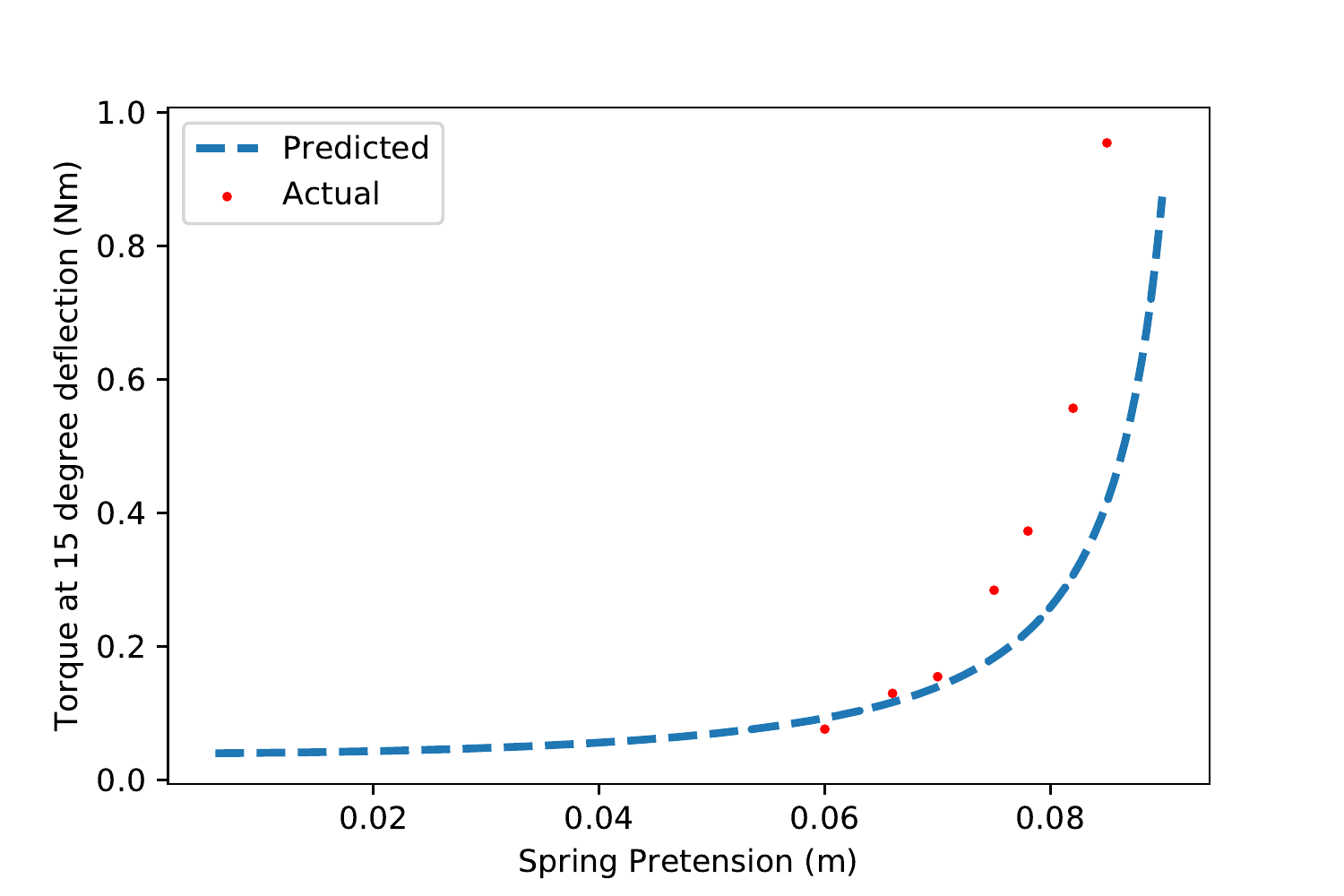}
    \caption{Predicted (blue dashed) and measured (red points) torque of actuator at different pretension. Torque measured at a joint deflection of $\jointinitial = 15$ degrees. 
    }
    \label{fig:predictedActualStiff}
\end{figure}

The output torque of the actuator is a function of both the pretension ($x_i$) and the deflection angle ($\theta$). 
We carried out two tests to characterize the behavior of the actuator and compare it to the theoretical models presented in \sref{section:methodA}.
We use a torsion spring with leg length $r_s = 0.05$ meters and torsional spring constant $\torsionk = 0.5 N/\theta$.
To compare the relationship between pretension and joint stiffness of the actuator with the theoretical equation \eqref{eq:stiffness}, we fixed the motor pulley in place and measured the joint torque at a deflection of 15 degrees at various pretensions. 
\figref{fig:predictedActualStiff} demonstrates how our measured values, fit by a curve, compare to the predicted model. The large increase in torque occurs at a lower pretension than expected.
This slight variation is not problematic for our application as the spring pair still exhibits controllable behavior within a pretension range.

\begin{figure}
    \centering
    \includegraphics[width=\linewidth]{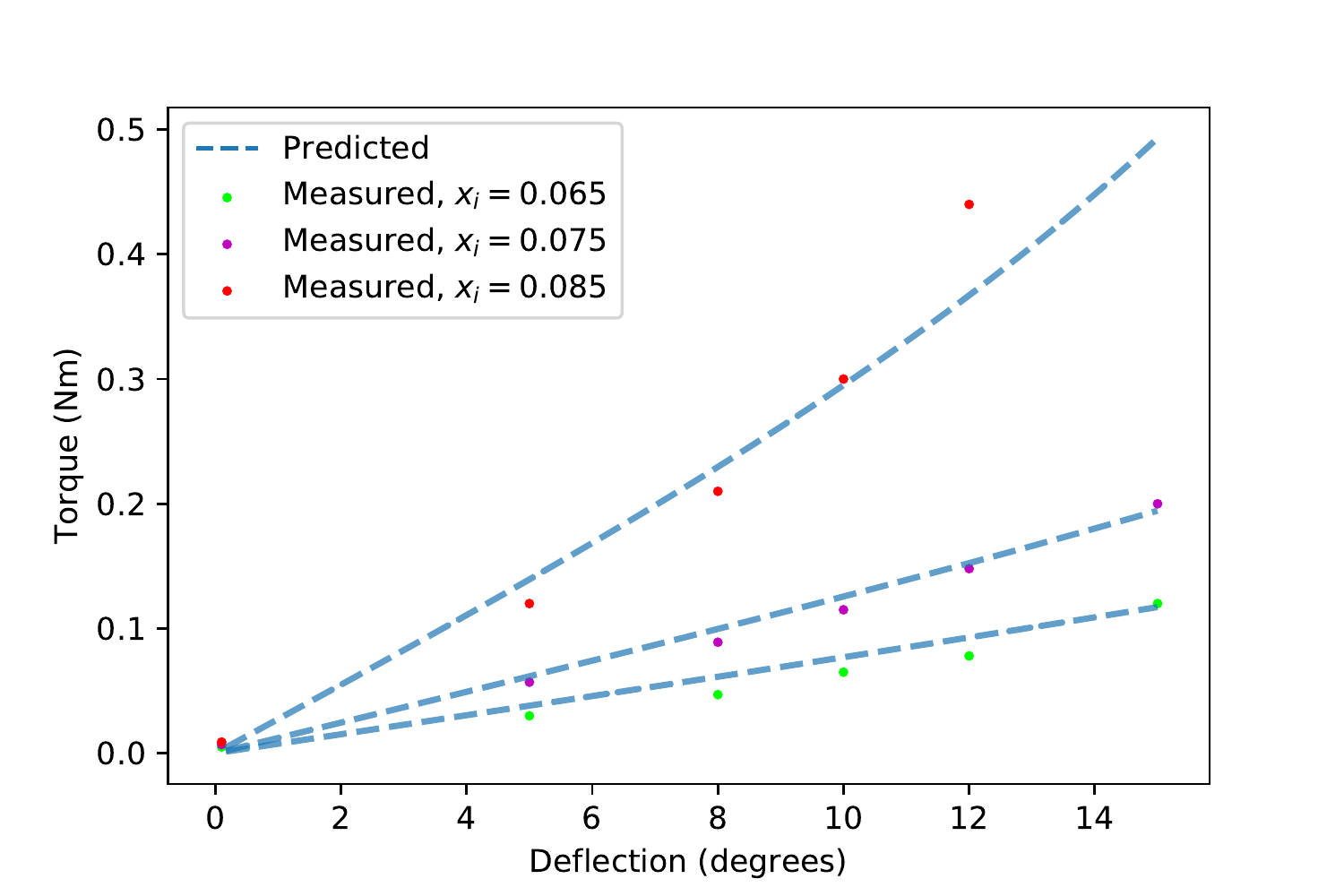}
    \caption{Torque vs. displacement of joint at pretension $\pretension = 0.065$m (green), $\pretension = 0.075$m (magenta), $\pretension = 0.085$m (red) . Plotted points shown over predicted curves (blue).
    }
    \label{fig:fixedTorque}
\end{figure}

Within a large range of pretension, the joint stiffness remains relatively the same, and quite low. 
At such a low stiffness, the actuator is unable to precisely control force or position, deeming the actuator useless. 
Additionally, designing a mechanism to extend the pretension from $0.00$ to $0.06$ meters would be wasteful, as the stiffness increases by a minute amount.
We therefore define the pretension range within which there is a significant change in stiffness as the operating stiffness. 
For the spring pair applied to our actuator, the operating stiffness is in the pretension range of  $\approx 0.06$m to $\approx 0.086$ meters.

We additionally fixed the pretension of the actuator at various values, and measured the torque vs. deflection relationship. 
\figref{fig:fixedTorque} displays the predicted relationship compared to the measured values, which are fit by a curve. 
We determine that the lower measured torque at a low deflection is likely due to the static friction force in the joint bearings, and the higher measured torque at a high deflection is the result of the inaccuracy observed in \figref{fig:predictedActualStiff}


\subsection{Hopping}
Our setup does not allow for the measurement of the forward velocity of the leg, so we instead experiment with hopping.
Hopping and running share many dynamic similarities as they are both cyclic motions that apply a force to the ground to accelerate the center of mass of the robot.

To repeatedly reverse the direction of the joint rotation, as is done in a hopping motion, a torque must be applied to decelerate the leg and then accelerate it in the opposite direction.
In a low stiffness setting, the springs must deflect a large amount from the motor position to achieve the required torque to rotate the leg. 
Once in motion, the springs will absorb the landing energy, and transfer it back into the leg motion as the cycle repeats. 
In a high stiffness setting, we predict that the rigid nature of the actuator will cause the leg to follow the motor position closely, with high stresses being placed on the structure of the leg as the leg reverses direction.

\begin{figure}
    \centering
    \includegraphics[width=0.9\linewidth]{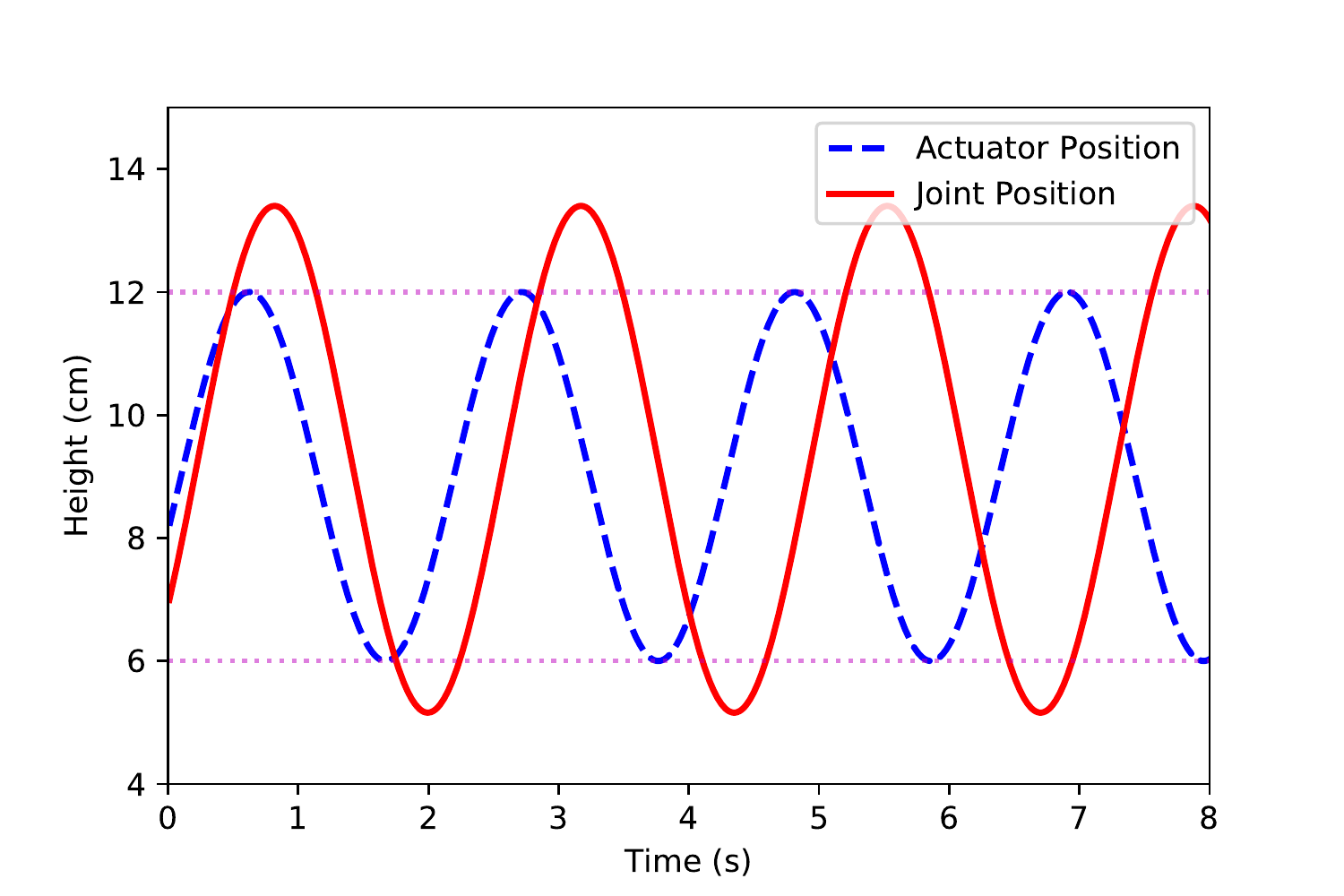}
    \caption{Hopping in low stiffness setting, with pretension $\pretension = 0.075$ meters. Blue dashed curve represents the motor position, assigned by the controller. The joint position is measured as the height from the ground measured by the ultrasonic sensor, and shown as the red curve.}
    \label{fig:softHop}
\end{figure}

We set the motor to position itself according to a periodic graph, and measured the position of the leg height and angle using an ultrasonic sensor and encoder. 
In a low stiffness setting, the joint position overshoots the motor position, and there is a higher period of oscillation (\figref{fig:softHop}). 
This demonstrates the low speed hopping associated with low stiffness.

\begin{figure}
    \centering
    \includegraphics[width=0.9\linewidth]{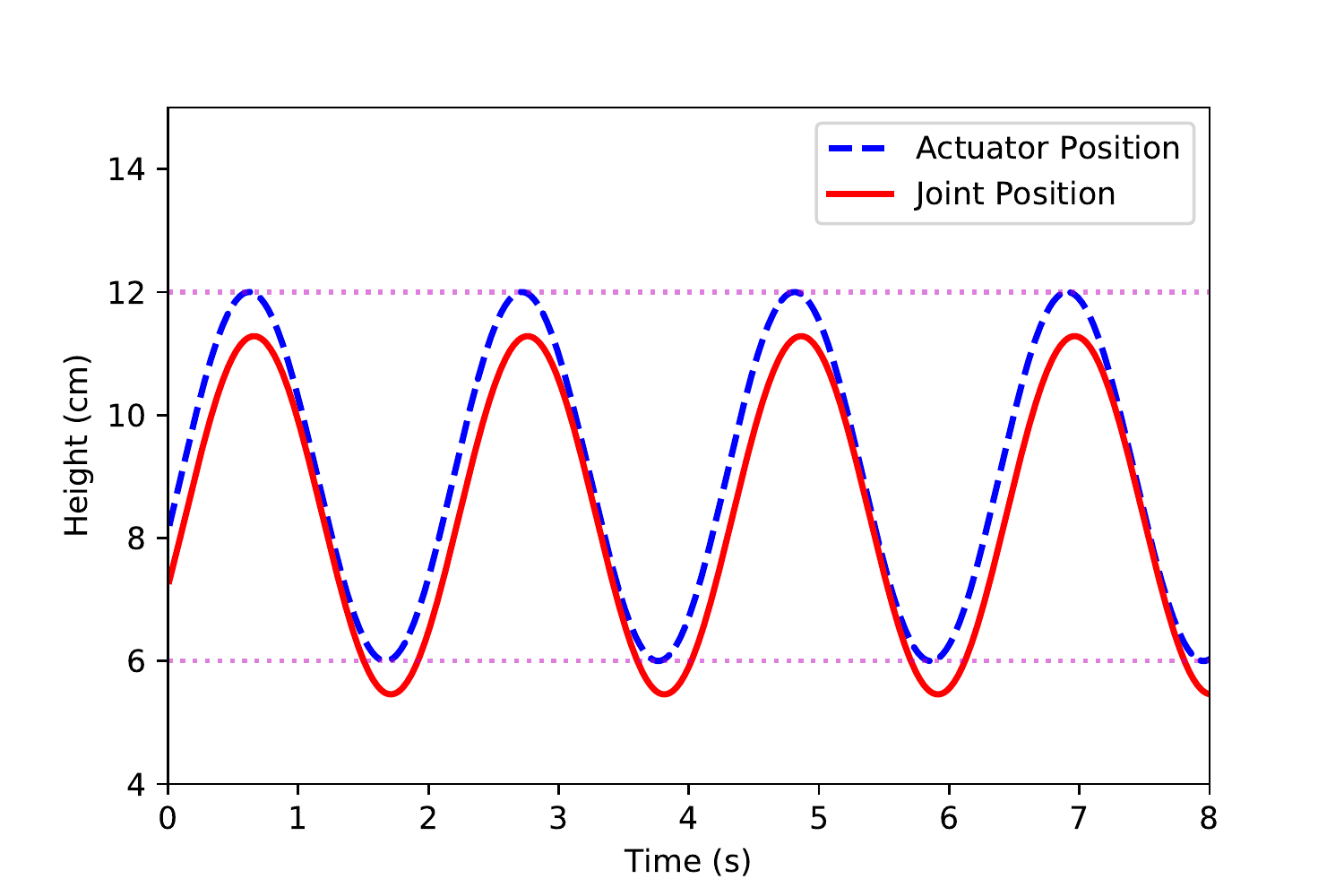}
    \caption{Hopping in high stiffness setting, with pretension $\pretension = 0.085$ meters. Targeted height curve from motor position shown in blue dashed line, and measured leg height using ultrasonic sensor represented by the red curve.}
    \label{fig:stiffHop}
\end{figure}

In a high stiffness setting, the system behaves similar to that of a non-elastic belt transmission, with the joint position aligning almost exactly with the actuator position (\figref{fig:stiffHop}).
We observed a slight offset, which is the result of the angular displacement required to create a torque that the spring-system must apply to balance the gravitational force. 
While compliance is desired in robotic legs for many applications, our experiments show that a high stiffness is required for high-speed hopping and running, making variable stiffness in legs necessary.

\subsection{Oscillation-less Drop}

We designed an experiment to find the optimal stiffness based on a set of parameters ($\groundheight, \dropheight + \secondheight, \jointinitial$) for which the leg does not oscillate upon landing, but rather absorbs the energy in it's springs. 
We set the final height of the system ($\groundheight$) to $29.5$cm, the drop distance ($\dropheight + \secondheight$) to $13.5$cm, and the initial leg angle $\jointinitial$ to $35$ degrees.
We modulate the stiffness by changing the pretension of the springs within the operating range defined in \sref{section: Characterization}.

We determine the theoretical pretension for this drop scenario to be $\pretension = 0.07746$m by allegorically iterating through the pretension values to find which value allows the leg to completely capture the gravitational potential energy in the springs, and have a net torque on the joint which balances the gravitational force at the desired final height $\groundheight$.

\begin{figure}
    \centering
    \includegraphics[width=\linewidth]{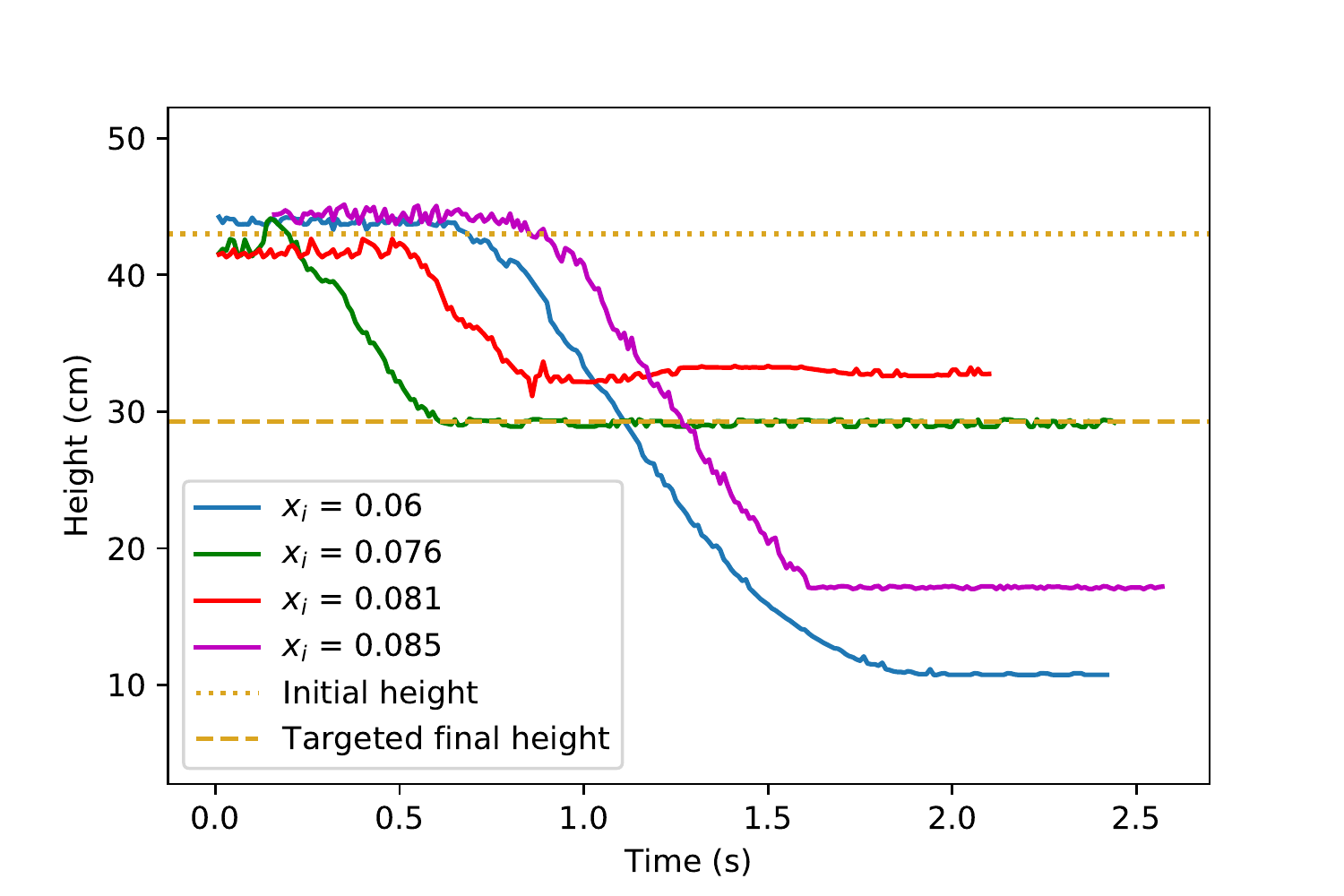}
    \caption{Dropping test with initial height of $\approx 43.5$cm, final height of $\approx 29$cm, and initial leg angle $\jointinitial = 35$ degrees. Tested at $\pretension = 0.06$m (blue), $\pretension = 0.076$m (green), $\pretension = 0.081$m (red), $\pretension = 0.086$m (magenta). The theoretical pretension was calculated to be $\pretension = 0.07746$m.}
    \label{fig: dropTest}
\end{figure}

We then tested multiple values to find the experimental optimal pretension.
\figref{fig: dropTest} displays that at a pretension of $x_i = 0.076$m, the desirable output is achieved. This compares well with our theoretical model, with an error of just $1.885 \%$. At this low error, the impact on the leg is minimal. We suspect that the measured pretension is lower as energy is lost to friction, and because lower torque from the springs is required as friction in the joints helps the leg remain upright. 

When dropping with a low stiffness, the torque on the joint is not high enough to overcome the gravitational force, and the leg drops to the ground. 
When dropping with a high stiffness, one of two different effects takes place. 
At a pretension of $x_i = 0.081$m, the joint is relatively stiff and oscillates slightly before arriving at rest at a final position greater than specified. 
At a pretension of $x_i = 0.085$m, the actuator experiences mechanical failure as the instantaneous force on the leg is great enough to cause the belt to skip on the motor pulley. 
The magenta plot demonstrates this, and shows that once the belt catches the pulley and stops skipping, the leg abruptly stops at a height below the desired final height.


%% file: sections/6_Discussion.tex
\section{Discussion}

The VSA design we propose in this paper has been shown to be mechanically simple and cheap. We demonstrate the large range of stiffness achieved at a low energy cost, with the ability to act analogous to a stiff actuator. Our experiments display the accuracy of our models, and effectiveness of our design.

While useful, our design has limitations. For a specific $\stiffnessparameter$ the torque $\jointtorque$ vs deflection $\deflection$ profile is fixed and may not be desirable in all situations. At a low stiffness, the belt often skips on the pulleys as the tension in the belt is not large enough to transmit power through the teeth. At a high stiffness, the belt sometimes skips on the pulley due to extremely high loads, as observed in the drop experiment. By placing the springs in the belt itself, we remove the ability for the joint to be continuously revolving, as past a certain deflection the spring will run into the output pulley.

Our next steps are to take the simplicity of our design into a more modular setup, which can revolve continuously and be applied to robots without belt transmissions. We also aim to design a passive VSA with a controllable torque vs deflection profile.


%% file: main.bbl
\begin{thebibliography}{10}
\providecommand{\url}[1]{#1}
\csname url@samestyle\endcsname
\providecommand{\newblock}{\relax}
\providecommand{\bibinfo}[2]{#2}
\providecommand{\BIBentrySTDinterwordspacing}{\spaceskip=0pt\relax}
\providecommand{\BIBentryALTinterwordstretchfactor}{4}
\providecommand{\BIBentryALTinterwordspacing}{\spaceskip=\fontdimen2\font plus
\BIBentryALTinterwordstretchfactor\fontdimen3\font minus
  \fontdimen4\font\relax}
\providecommand{\BIBforeignlanguage}[2]{{%
\expandafter\ifx\csname l@#1\endcsname\relax
\typeout{** WARNING: IEEEtran.bst: No hyphenation pattern has been}%
\typeout{** loaded for the language `#1'. Using the pattern for}%
\typeout{** the default language instead.}%
\else
\language=\csname l@#1\endcsname
\fi
#2}}
\providecommand{\BIBdecl}{\relax}
\BIBdecl

\bibitem{Zinn2004playing}
M.~Zinn, O.~Khatib, B.~Roth, and J.~Salisbury, ``Playing it safe
  [human-friendly robots],'' \emph{IEEE Robotics Automation Magazine}, vol.~11,
  no.~2, pp. 12--21, 2004.

\bibitem{an2019mechanical}
Z.~An and C.~Li, ``Mechanical simulation and full order sliding mode collision
  avoidance compliant control based on neural network of dual-arm space robot
  with compliant mechanism capturing satellite 1,'' \emph{Chinese Journal of
  Theoretical and Applied Mechanics}, vol.~51, no.~4, p. 1156, 2019.

\bibitem{Hu2014compliance}
Y.~Hu, M.~Felis, and K.~Mombaur, ``Compliance analysis of human leg joints in
  level ground walking with an optimal control approach,'' in \emph{2014
  IEEE-RAS International Conference on Humanoid Robots}, 2014, pp. 881--886.

\bibitem{Villani2016force}
L.~Villani and J.~De~Schutter, ``Force control,'' in \emph{Springer handbook of
  robotics}.\hskip 1em plus 0.5em minus 0.4em\relax Springer, 2016, pp.
  195--220.

\bibitem{Ham2009compliant}
V.~Ham, T.~Sugar, B.~Vanderborght, K.~Hollander, and D.~Lefeber,
  ``\BIBforeignlanguage{English (US)}{Compliant actuator designs: Review of
  actuators with passive adjustable compliance/controllable stiffness for
  robotic applications},'' \emph{\BIBforeignlanguage{English (US)}{IEEE
  Robotics and Automation Magazine}}, vol.~16, no.~3, pp. 81--94, 2009,
  copyright: Copyright 2009 Elsevier B.V., All rights reserved.

\bibitem{rummel2008stable}
J.~Rummel and A.~Seyfarth, ``Stable running with segmented legs,'' \emph{The
  International Journal of Robotics Research}, vol.~27, no.~8, pp. 919--934,
  2008.

\bibitem{farley1991hopping}
C.~T. Farley, R.~Blickhan, J.~Saito, and C.~R. Taylor, ``Hopping frequency in
  humans: a test of how springs set stride frequency in bouncing gaits,''
  \emph{Journal of applied physiology}, vol.~71, no.~6, pp. 2127--2132, 1991.

\bibitem{mutlu2018effects}
M.~Mutlu, S.~Hauser, A.~Bernardino, and A.~J. Ijspeert, ``Effects of passive
  and active joint compliance in quadrupedal locomotion,'' \emph{Advanced
  Robotics}, vol.~32, no.~15, pp. 809--824, 2018.

\bibitem{galloway2010}
\BIBentryALTinterwordspacing
K.~C. Galloway, ``Passive variable compliance for dynamic legged robots,''
  \emph{Publicly Accessible Penn Dissertations. 246}, 2010. [Online].
  Available: \url{http://repository.upenn.edu/edissertations/246}
\BIBentrySTDinterwordspacing

\bibitem{wolf2015variable}
S.~Wolf, G.~Grioli, O.~Eiberger, W.~Friedl, M.~Grebenstein, H.~H{\"o}ppner,
  E.~Burdet, D.~G. Caldwell, R.~Carloni, M.~G. Catalano \emph{et~al.},
  ``Variable stiffness actuators: Review on design and components,''
  \emph{IEEE/ASME transactions on mechatronics}, vol.~21, no.~5, pp.
  2418--2430, 2015.

\bibitem{hurst2008role}
J.~W. Hurst, ``The role and implementation of compliance in legged
  locomotion,'' Ph.D. dissertation, Carnegie Mellon University, The Robotics
  Institute, 2008.

\bibitem{Pratt1995series}
G.~Pratt and M.~Williamson, ``Series elastic actuators,'' in \emph{Proceedings
  1995 IEEE/RSJ International Conference on Intelligent Robots and Systems.
  Human Robot Interaction and Cooperative Robots}, vol.~1, 1995, pp. 399--406
  vol.1.

\bibitem{English1999mechanics}
\BIBentryALTinterwordspacing
C.~English and D.~Russell, ``Mechanics and stiffness limitations of a variable
  stiffness actuator for use in prosthetic limbs,'' \emph{Mechanism and Machine
  Theory}, vol.~34, no.~1, pp. 7--25, 1999. [Online]. Available:
  \url{https://www.sciencedirect.com/science/article/pii/S0094114X98000263}
\BIBentrySTDinterwordspacing

\bibitem{Tonietti2005design}
G.~Tonietti, R.~Schiavi, and A.~Bicchi, ``Design and control of a variable
  stiffness actuator for safe and fast physical human/robot interaction,'' in
  \emph{Proceedings of the 2005 IEEE International Conference on Robotics and
  Automation}, 2005, pp. 526--531.

\bibitem{Migliore2005biologically}
S.~Migliore, E.~Brown, and S.~DeWeerth, ``Biologically inspired joint stiffness
  control,'' in \emph{Proceedings of the 2005 IEEE International Conference on
  Robotics and Automation}, 2005, pp. 4508--4513.

\bibitem{catalano2011VSAcube}
M.~G. Catalano, G.~Grioli, M.~Garabini, F.~Bonomo, M.~Mancini, N.~Tsagarakis,
  and A.~Bicchi, ``Vsa-cubebot: A modular variable stiffness platform for
  multiple degrees of freedom robots,'' in \emph{2011 IEEE International
  Conference on Robotics and Automation}, 2011, pp. 5090--5095.

\bibitem{Hurst2004AMASC}
J.~Hurst, J.~Chestnutt, and A.~Rizzi, ``An actuator with physically variable
  stiffness for highly dynamic legged locomotion.'' vol. 2004, 01 2004, pp.
  4662--4667.

\bibitem{chou1996measurement}
C.-P. Chou and B.~Hannaford, ``Measurement and modeling of mckibben pneumatic
  artificial muscles,'' \emph{IEEE Transactions on robotics and automation},
  vol.~12, no.~1, pp. 90--102, 1996.

\bibitem{raibert1989dynamically}
M.~H. Raibert, H.~B. Brown~Jr, M.~Chepponis, J.~Koechling, and J.~K. Hodgins,
  ``Dynamically stable legged locomotion,'' Massachusetts Inst of Tech
  Cambridge Artificial Intelligence Lab, Tech. Rep., 1989.

\bibitem{Hollander2005Jack}
K.~Hollander, T.~Sugar, and D.~Herring, ``Adjustable robotic tendon using a
  'jack spring'/spl trade/,'' in \emph{9th International Conference on
  Rehabilitation Robotics, 2005. ICORR 2005.}, 2005, pp. 113--118.

\bibitem{Morita1995DesignAD}
T.~Morita and S.~Sugano, ``Design and development of a new robot joint using a
  mechanical impedance adjuster,'' \emph{Proceedings of 1995 IEEE International
  Conference on Robotics and Automation}, vol.~3, pp. 2469--2475 vol.3, 1995.

\bibitem{hollander2004concepts}
K.~Hollander and T.~Sugar, ``Concepts for compliant actuation in wearable
  robotic systems.''

\bibitem{geeroms2018energetic}
J.~Geeroms, L.~Flynn, R.~Jimenez-Fabian, B.~Vanderborght, and D.~Lefeber,
  ``Energetic analysis and optimization of a maccepa actuator in an ankle
  prosthesis,'' \emph{Autonomous Robots}, vol.~42, no.~1, pp. 147--158, 2018.

\bibitem{duindam2005optimization}
V.~Duindam and S.~Stramigioli, ``Optimization of mass and stiffness
  distribution for efficient bipedal walking,'' in \emph{Proceedings of the
  International Symposium on Nonlinear Theory and Its Applications}.\hskip 1em
  plus 0.5em minus 0.4em\relax Citeseer, 2005.

\bibitem{daskalov1990kinematic}
A.~Y. Daskalov, ``Kinematic analysis of cardan drives,'' \emph{Mechanism and
  machine theory}, vol.~25, no.~5, pp. 479--486, 1990.

\end{thebibliography}
